\documentclass[10pt,journal,compsoc]{IEEEtran}

\ifCLASSOPTIONcompsoc
  \usepackage[nocompress]{cite}
\else
  \usepackage{cite}
\fi

\ifCLASSINFOpdf
  \usepackage[pdftex]{graphicx}
  \graphicspath{{images/}}
  \DeclareGraphicsExtensions{.pdf,.jpeg,.jpg,.png}
\else
\fi

\usepackage{amsmath}
\usepackage{array}
\usepackage{booktabs}

\usepackage{url}

\usepackage{tikz}
\newcommand{\ditto}[1]{%
    \tikz{
        \draw [line width=0.12ex] (-0.2ex,0) -- +(0,0.8ex)
            (0.2ex,0) -- +(0,0.8ex);
        \draw [line width=0.08ex] (-0.6ex,0.4ex) -- +(-#1em, 0)
            (0.6ex,0.4ex) -- +(#1em,0);
    }%
}

\begin{document}

\title{Utilizing Neural Networks and Linguistic Metadata for Early Detection of Depression Indications in Text Sequences}

\author{Marcel~Trotzek, Sven~Koitka, and~Christoph~M.~Friedrich,~\IEEEmembership{Member,~IEEE}
\IEEEcompsocitemizethanks{\IEEEcompsocthanksitem The authors are with the Department
of Computer Science, University of Applied Sciences and Arts Dortmund, Germany.\protect\\
E-mail: mtrotzek@stud.fh-dortmund.de, sven.koitka@fh-dortmund.de, and christoph.friedrich@fh-dortmund.de
\IEEEcompsocthanksitem S. Koitka is also with the Department of Computer Science, TU Dortmund University, Germany, and with the Department of Diagnostic and Interventional Radiology and Neuroradiology, University Hospital Essen, Germany.
\IEEEcompsocthanksitem C. M. Friedrich is also with the Institute for Medical Informatics, Biometry and Epidemiology (IMIBE), University Hospital Essen, Germany.}

\thanks{This work has been submitted to the IEEE and has been accepted for future publication. Copyright may be transferred without notice, after which this version may no longer be accessible.}}

\markboth{Accepted for publication in a future issue of IEEE Transactions on Knowledge and Data Engineering: DOI 10.1109/TKDE.2018.2885515\hspace{0.8em}}%
{Trotzek \MakeLowercase{\textit{et al.}}: Utilizing Neural Networks and Linguistic Metadata for Early Detection of Depression Indications in Text Sequences}

\IEEEtitleabstractindextext{%
\begin{abstract}
Depression is ranked as the largest contributor to global disability and is also a major reason for suicide. Still, many individuals suffering from forms of depression are not treated for various reasons. Previous studies have shown that depression also has an effect on language usage and that many depressed individuals use social media platforms or the internet in general to get information or discuss their problems. This paper addresses the early detection of depression using machine learning models based on messages on a social platform. In particular, a convolutional neural network based on different word embeddings is evaluated and compared to a classification based on user-level linguistic metadata. An ensemble of both approaches is shown to achieve state-of-the-art results in a current early detection task. Furthermore, the currently popular $ERDE$ score as metric for early detection systems is examined in detail and its drawbacks in the context of shared tasks are illustrated. A slightly modified metric is proposed and compared to the original score. Finally, a new word embedding was trained on a large corpus of the same domain as the described task and is evaluated as well.
\end{abstract}

}

\maketitle

\IEEEdisplaynontitleabstractindextext
\IEEEpeerreviewmaketitle

\IEEEraisesectionheading{\section{Introduction}\label{sec:introduction}}
\IEEEPARstart{A}{ccording} to World Health Organization (WHO)\cite{who2017depression}, more than 300 million people worldwide are suffering from depression, which equals about 4.4\% of the global population. While forms of depression are more common among females (5.1\%) than males (3.6\%) and prevalence differs between regions of the world, it occurs in any age group and is not limited to any specific life situation. Depression is therefore often described to be accompanied by paradoxes, caused by a contrast between the self-image of a depressed person and the actual facts\cite{beck2009depression}. Latest results from the 2016 National Survey on Drug Use and Health in the United States\cite{NSDUH2017summary} report that, during the year 2016, 12.8\% of adolescents between 12 and 17 years old and 6.7\% of adults had suffered a major depressive episode (MDE).

Precisely defining depression is not an easy task, not only because several sub-types have been described and changed in the past\cite{paykel2008basic}, but also because the term ``being depressed'' has become frequently used in everyday language. In general, depression can be described to lead to an altered mood and may also be accompanied, for example, by a negative self-image, wishes to escape or hide, vegetative changes, and a lowered overall activity level\cite[p. 8]{beck2009depression}. The symptoms experienced by depressed individuals can severely impact their ability to cope with any situation in daily life and therefore differ drastically from normal mood variations that anyone experiences.

At the worst, depression can lead to suicide. WHO estimates that, in the year 2015, 788,000 people have died by suicide and that it was the second most common cause of death for people between 15 and 29 years old worldwide\cite{who2017depression}. In Europe, self-harm was even reported as the most common cause of death in the age group between 15 and 29 and the second most common between 30 and 49, again in results obtained by WHO in 2015\cite{who2016deaths}.

Although the severity of depression is well-known, only about half of the individuals affected by any mental disorder in Europe get treated\cite{alonso2007population}. The proportion of individuals seeking treatment for mood disorders during the first year ranges between 29--52\% in Europe, 35\% in the USA, and only 6\% in Nigeria or China\cite{wang2007delay}. In addition to possible personal reasons for avoiding treatment, this is often due to a limited availability of mental health care, for example in conflict regions\cite{rahman2016effect}. Via a telephone survey in Germany\cite{schomerus2009stigma}, researchers found out that shame and self-stigmatization seem to be much stronger reasons to not seek psychiatric help than actual perceived stigma and negative reactions of others. They further speculate that the fear of discrimination might be relatively unimportant in their study because people hope to keep their psychiatric treatment secret. Another study amongst people with severe mental illness in Washington D.C. showed that stigma and discrimination indeed exist, while they are not ``commonly \emph{experienced} problems'' but rather ``perceived as omnipresent \emph{potential} problems''\cite[p. 1]{whitley2014stigma}.

While depression and other mental illnesses may lead to social withdrawal and isolation, it was found that social media platforms are indeed increasingly used by affected individuals to connect with others, share experiences, and support each other\cite{gowen2012young,naslund2014naturally}. Based on these findings, peer-to-peer communities on social media can be able to challenge stigma, increase the likelihood to seek professional help, and directly offer help online to people with mental illness\cite{naslund2016future}. A similar study in the USA\cite{berger2005internet} came to the conclusion that internet users with stigmatized illnesses like depression or urinary incontinence are more likely to use online resources for health-related information and for communication about their illness than people with another chronic illness. All this emphasizes the importance of research toward ways to assist depressed individuals on social media platforms and on the internet in general.

This paper is therefore focused on ways to classify indications of depression in written texts as early as possible based on machine learning methods. The work presented in this paper is structured as follows: Section \ref{chap:related} gives an overview of related work concerning depression, its influence on language, and natural language processing methods. Section \ref{chap:dataset} describes the dataset used in this work, analyzes the evaluation metric of the corresponding task, and proposes an alternative. Section \ref{chap:metadata} introduces the user-based metadata features used for classification, while Section \ref{chap:network} describes the neural network models utilized for this task. Section \ref{chap:experiments} contains an experimental evaluation of these models and compares them to published results. Finally, Section \ref{chap:conclusion} concludes this work and summarizes the results.

\section{Related Work}
\label{chap:related}
This section describes the context of this work based on previous research concerning depression and its effects on language. Since social media research in general and health research in particular require ethical considerations, an overview of the current ethical discussion in the field of natural language processing is given. Finally, the practical basis of this work is described by investigating previous and current work in text classification using machine learning.

\subsection{Depression and Language}
\label{sec:language}
Previous studies have already shown that depression also has an effect on the language used by affected individuals. For example, a more frequent use of first person singular pronouns in spoken language was first observed in 1981\cite{bucci1981language,weintraub1981verbal}. An examination of essays written by depressed, formerly-depressed, and non-depressed college students at University of Texas\cite{rude2004language} confirmed an elevated use of the word ``I'' in particular and also found more negative emotion words in the depressed group. Similarly, a Russian speech study\cite{smirnova2013language} found a more frequent use of all pronouns and verbs in past tense among depression patients. A recent study based on English forum posts\cite{almosaiwi2018absolutist} observed an elevated use of absolutist words (e.g. absolutely, completely, every, nothing\footnote{A full list of words is available as Table S2 from \url{http://journals.sagepub.com/doi/suppl/10.1177/2167702617747074}, accessed on 2018-02-14}) within forums related to depression, anxiety, and suicidal ideation than within completely unrelated forums as well as ones about asthma, diabetes, or cancer.

The knowledge that language can be an indicator of an individual's psychological state has, for example, lead to the development of the Linguistic Inquiry and Word Count (LIWC) software\cite{tausczik2010psychological,pennebaker2003psychological}. By utilizing  a comprehensive dictionary, it allows researchers to evaluate written texts in several categories based on word counts. A more detailed description of LIWC and its features is given in Section \ref{chap:metadata}. With a similar purpose, Differential Language Analysis Toolkit (DLATK)\cite{schwartz2017dlatk}, an open-source Python library, was created for text analysis with a psychological, health, or social focus.

\subsection{Ethical Perspective}
Driven by the growing availability of data, for example through social media, and the technological advances that allow researchers to work with this data, ethical considerations are becoming more and more important in the field of Natural Language Processing (NLP). Based on these developments, NLP has changed from being mostly focussed on improving linguistic analysis towards actually having an impact on individuals based on their writings. Still, a proper discussion about ethics in NLP has only been started in 2016 by Hovy and Spruit\cite{hovy2016social}. Although Institutional Review Boards (IRBs) have been well-established to enforce ethical guidelines on experiments that directly involve human subjects, the authors note that NLP and data sciences in general have not constructed such guidelines. They further argue that language ``is a \textit{proxy for human behavior}, and a strong signal of individual characteristics'' and that, in addition, ``the texts we use in NLP carry \textit{latent} information about the author and situation''\cite[p. 592]{hovy2016social}. On top of this direct connection to the individual, they also describe the social impact of NLP research\cite[pp. 593--594]{hovy2016social}. A demographic bias in the selection of training texts can lead to the \textit{exclusion} of specific groups, \textit{overgeneralization} based on false positives can have serious consequences depending on the task, and research results can potentially cause or confirm biases and ultimately discrimination by \textit{topic overexposure}. Even if all these factors are considered, they conclude that \textit{dual-use problems} can exist for any research if results are used in a different way than originally intended. The same applies to pre-trained machine learning models that get published and could theoretically be used in unintended ways.

These discussions about ethics in NLP have lead to the First Workshop on Ethics in Natural Language Processing\footnote{\url{http://www.ethicsinnlp.org/}, accessed on 2018-02-14} during the conference of the European Chapter of the Association for Computational Linguistics in 2017 (EACL 2017). Some interesting results of this workshop include, for example, a proposed process to make NLP research ``ethical by design''\cite{leidner2017ethical} by installing an Ethics Review Board (ERB) in research organizations that has to approve or veto all steps during research, development, and deployment. Specifically for health research in social media, guidelines for ethical research have been proposed\cite{benton2017ethical}. They include obtaining consent from users whenever possible, carefully considering the consequences of any interactions with users or modifications of the user experience, protecting the data during research and when sharing it with other researchers, and de-identifying users during analysis, presentation, and when linking data from several platforms. From another perspective, there are also ethical considerations to keep in mind for NLP shared tasks and shared tasks in general\cite{escartin2017ethical}. The competitive nature of such tasks may lead researchers to be secretive about their systems and methods, ethical concerns may be overlooked, and conflicts of interest may arise if organizers themselves participate in a task. 

While most discussions about health research in social media focus on the important theoretical groundwork to establish guidelines, there has also been a qualitative study using focus group interviews with 16 depressed and 10 non-depressed participants\cite{mikal2016ethical} to investigate their opinion about population-level mental health monitoring on Twitter. Firstly, participants of this study were generally aware of the fact that their Twitter messages are public, but showed misconceptions about how access to them could be limited by deleting them, by limitations of the user interface, or by the sheer amount of messages on the platform. While the participants mainly accepted aggregated depression monitoring based on Twitter, some still found it ``creepy'' and a particular participant stated: ``The fact that if it was an algorithm, and they were looking like, `Hey, we think you're feeling low right now.' I feel like it might make me feel even more low.''\cite[p. 6]{mikal2016ethical} Similar to this statement, participants were concerned about the possibility to use population-based data to identify specific individuals, while others had the opinion that ``pinpointing individuals could help them access much-needed mental health services by paying attention to cues that friends may ignore''\cite[p. 7]{mikal2016ethical}. In general, participants supported the idea to use social media data as an additional source for professional therapists.

\subsection{Natural Language Processing}
The work described in this paper belongs to the area of Natural Language Processing (NLP)\cite{manning1999foundations} and text classification in particular. The origins of text classification tasks can be found in early research to automatically categorize documents based on statistical analysis of specific clue words in 1961\cite{maron1961automatic}. Later, similar research goals lead to rule-based text classification systems  like CONSTRUE in 1990\cite{hayes1990tcs} and finally the field began to shift more and more to machine learning algorithms around the year 2000\cite{mitchell1997machine,sebastiani2002machine}. In addition to text categorization, machine learning was also a driving force in other text-based tasks like sentiment analysis, which is focussed on extracting opinions and sentiment from text documents\cite{pang2008opinion}. It was first used in combination with machine learning to find positive or negative opinions in movie reviews \cite{pang2002thumbs} and was then extended to other review domains\cite{turney2002thumbs}, as well as completely different areas like social media monitoring and general analysis of consumer attitudes\cite{pang2008opinion}.

More recently, deep learning has been utilized for text classification\cite{johnson2017deep, peters2018deep} in addition to its more common usages in image classification. State-of-the-art results in several text-based tasks could, for example, be achieved by transfer learning methods like Universal Language Model Fine-tuning (ULMFit)\cite{howard2018universal} and the Google research project Bidirectional Encoder Representations from Transformers (BERT)\cite{devlin2018bert} for the training of language representations, which includes ULMFit and several other methods. The code of BERT and several pre-trained models are also available on GitHub\footnote{\url{https://github.com/google-research/bert}, accessed on 2018-11-24}.

Based on these developments, research evolved to text classification tasks that extract more than just opinions from documents: Especially the availability of social media messages enabled researchers to extract population-based health information that made it possible to track diseases, symptoms, and medications\cite{paul2011you}. More specifically, Twitter messages were used for population level tracking of depression\cite{choudhury2013social}, detection of depression\cite{choudhury2013predicting,nadeem2016identifying}, bipolar disorder\cite{huang2017detection}, and post traumatic stress disorder (PTSD)\cite{coppersmith2015clpsych} for individuals. Depression detection from text documents in particular has become an increasingly important research area, with interesting methods and results reported for Twitter, Facebook, and forum posts\cite{guntuku2017detecting}. To directly help depression patients, systems like Psychologist in a Pocket\cite{bitsch2015piap, cheng2016psychologist}, an Android smartphone app, are being developed: Users of this app can choose specific text inputs on their device that should be monitored (e.g. social media posts, mails, or text messages) to be informed about possibly alarming mood changes that they themselves might overlook. By installing an additional plugin, data can be shared with a third party, for example a therapist, and is otherwise password secured and only saved locally.

In addition to text-based depression detection, the second sub-task of the work described in this paper can be found in the area of early detection. Early detection based on text documents can be seen to originate from the idea of sequential reading to allow predictions based on as few documents as possible\cite{amold2011text}. An approach using a modified na\"ive Bayes classifier was shown to be viable for text categorization and sexual predator detection with partial information\cite{escalante2016early}. Other interesting use cases of early detection applied in practice have been found in the detection of early signs of epidemics\cite{torii2011exploratory} or rumors\cite{zhao2015enquiring} from social media messages.

The fields of depression detection and early detection were first combined by the publication of a dataset for early detection of depression in reddit messages\cite{losada2016test} and research using this dataset was driven by the Conference and Labs of the Evaluation Forum (CLEF) 2017 conference\footnote{\url{http://clef2017.clef-initiative.eu/}, accessed on 2018-02-18} workshop on early risk detection on the internet\cite{losada2017erisk, losada2017lncs}. As this task and dataset are also utilized in this paper, further details can be found in Section \ref{chap:dataset}. During the workshop in 2017, interesting results could be obtained using combinations of Information Retrieval (IR) and supervised learning based on bag of words and dictionaries\cite{almeida2017erisk}, a two-step classification based first on posts and then on users\cite{anzaldua2017erisk}, purely user-based features and random forests\cite{malam2017erisk}, lexicon word counts and medial concepts using Support Vector Machines (SVM) or Recurrent Neural Networks (RNN) with Gated Recurrent Units (GRU)\cite{sadeque2017erisk}, and graph models\cite{tello2017erisk}. The Temporal Variation of Terms (TVT) model for early detection, based on the variation of vocabulary over time, was proposed\cite{errecalde2017erisk} and successfully evaluated\cite{villegas2017erisk}. The authors of this paper participated in the task by using models that combined user-based linguistic metadata with bag of words, document embeddings, and RNNs using Long Short Term Memory (LSTM) layers\cite{trotzek2017erisk}. Results from this task are used to evaluate the experiments in Section \ref{chap:experiments}.

Similar text classification research in a psychological context has been conducted at the CLPsych conferences\footnote{\url{http://clpsych.org/}, accessed on 2018-11-24} of the past years. In 2016 and 2017, for example, the conference presented a shared task\cite{milne2016clpsych, milne2017clpsych} that challenged participants to prioritize posts in an online peer-support forum to tell moderators how urgently a message needs their attention. The CLPsych shared task in 2018\cite{W18-0604} focused on an even more notable approach to early detection: Based on essays written by 11-year-olds, participants had to predict the current as well as future psychological state of the author at specific times in their life.

\section{Dataset Overview}
\label{chap:dataset}
This section gives an overview of the dataset used for the experiments described in this paper and its main characteristics. It also details the corresponding task and the evaluation criteria.

\subsection{Dataset}
\label{sec:dataset}
The dataset utilized in all experiments for this paper was first described in 2016 for research on depression and language use\cite{losada2016test} and then finally published as part of the CLEF 2017 conference eRisk pilot task on early detection of depression\cite{losada2017erisk}. It contains chronological sequences of posts and comments from \url{reddit.com}, collected for a total of 135 depressed users and a random control group of 752 users. Depressed users were identified by searching for posts that clearly mention a diagnosis (e.g. ``I was diagnosed with depression''). Since there is no way to validate these statements and no further investigation of the users was possible, there could theoretically be non-depressed individuals in this group but also depressed ones in the control group. Any occurrence of user names has been replaced by an ID like \textit{train\_subject\_1} to anonymize users. The number of messages collected for each user ranges from 10 to 2,000 due to API limitations and the fact that some of them have posted very rarely. The dataset has been split into a training and test set as displayed in Table \ref{tab:dataset}.

\begin{table}[!t]
	\renewcommand{\arraystretch}{1.3}
	\caption{Main Statistics of the eRisk 2017 Dataset. Adapted from \cite{losada2017erisk} and Extended.}
	\label{tab:dataset}
	\centering
	\begin{tabular}{lcccc}
		\toprule
		 & \multicolumn{2}{c}{Train} & \multicolumn{2}{c}{Test}\\
		 & \emph{Depressed} & \emph{Control} & \emph{Depressed} & \emph{Control}\\
		\midrule
		Users & 83 & 403 & 52 & 349\\
		Messages & 30,851 & 264,172 & 18,706 & 217,665\\
		\hspace*{10pt}Links/Title only & 2,768 & 81,474 & 973 & 56,543\\
		\hspace*{10pt}Title + Text & 2,143 & 9,907 & 955 & 9,192\\
		\hspace*{10pt}Comments & 25,939 & 172,746 & 16,776 & 151,887\\
		\hspace*{10pt}Empty messages & 1 & 45 & 2 & 43\\
		Avg. msgs. per user & 371.7 & 655.5 & 359.7 & 623.7\\
		\bottomrule
	\end{tabular}
\end{table}

Each message in the dataset may consist of a \emph{title}, \emph{text}, or both, depending on its type: Users on reddit are able to post content in terms of an image or URL (\emph{title} only), as text content (\emph{title} and optional \emph{text}), or as comment on another message (\emph{text} only). A total of 91 messages in the dataset are completely empty and can therefore be discarded. Since deleted messages are normally exchanged with the text ``[deleted]'', these seem to be caused by a fault in reddit, the API, or the preprocessing before publishing the dataset.

In addition, each message also contains a \emph{date} attribute with the timestamp of when the user has published it exact to the second. Since the reddit API returns all timestamps in UTC (or the local timezone of the reddit server\footnote{\url{https://github.com/praw-dev/praw/issues/243}, accessed on 2018-01-24}), these timestamps can primarily be used to sort messages and search for time patterns of a single user. Comparing timestamps between different users would most likely give misleading results because their actual timezone is unknown and they could live anywhere in the world.

Since users for the control group were collected by selecting users that had posted recently when the dataset was collected, instead of using a distribution over time similar to the depressed users, the timestamps also contain a hidden feature that could be exploited: When using the time of the latest post per user (in seconds since epoch) as only input for a logistic regression, this single feature was enough to obtain an $F_1$ score of $0.78$ on the test data. This feature could easily be used as soon as the last data chunk (see Section \ref{sec:task}) is available. As this is clearly not intended and not in the interest of this task, all models created for this paper completely discard the timestamp information and a detailed analysis of this fact has been sent to the organizers of eRisk to prevent this in future tasks\footnote{According to the organizers, this will already be done for the eRisk 2018 task.}.

\subsection{Task and Evaluation Criteria}
\label{sec:task}
The given dataset was explicitly published for research toward early detection of depression within the previously described eRisk task. To measure this criterion, the data was also split into ten chunks by the organizers, containing 10\% of each user's messages in chronological order. During the test phase of the eRisk task, a single chunk of data was published each week, starting with the oldest messages of the users. Participants then had the possibility to classify a user as depressed, non-depressed, or delay the decision to see additional data in the next week. Submitted predictions were final and could not be reversed later. In the last week, a prediction had to be given for every user. In addition to the correct and wrong predictions, evaluations could therefore also take into account how many messages participants had seen for each user before giving a prediction. This information can be utilized by the organizers' early risk detection error (ERDE) measure for early detection systems that was defined in their dataset paper as well\cite[pp. 7--8]{losada2016test}: With a binary decision $d$ submitted for a user after reading $k$ of his messages, $ERDE_o$ is defined as:
\begin{equation}
	ERDE_o\left(d, k\right) =
	\begin{cases}
		c_{fp} & \text{for false positives (FP)}\\
		c_{fn} & \text{for false negatives (FN)}\\
		lc_{o}(k) \cdot c_{tp} & \text{for true positives (TP)}\\
		0 & \text{for true negatives (TN)}\,.
	\end{cases}
\end{equation}
The values of $c_{fp}$ and $c_{fn}$ can be used to adjust the severity of false positives and false negatives to the given domain, while $c_{tp}$ defines how late predictions of positive cases are punished. For the eRisk 2017 task, $c_{fn}$ was set to $1$, $c_{fp}$ to $\dfrac{n_{tp}}{n_u}$, with $n_{tp}$ denoting the number of positive cases in the test data and the total number of test users $n_u$. Finally, $c_{tp}$ was set to 1 in order to treat late predictions equally to no prediction at all\cite[p. 5]{losada2017erisk}. The function $lc_o(k)$ determines after how many messages $k$ the cost for true positives starts to grow and is defined as:
\begin{equation}
	lc_{o}(k) = 1 - \dfrac{1}{1 + e^{k - o}}\,,
\end{equation}
where the free parameter $o$ controls around which point this logistic sigmoid function is centered. Results of the eRisk 2017 task were evaluated based on $ERDE_{5}$, $ERDE_{50}$, and $F_1$ score.

\begin{figure}[!t]
	\centering
	\includegraphics[width=3.5in]{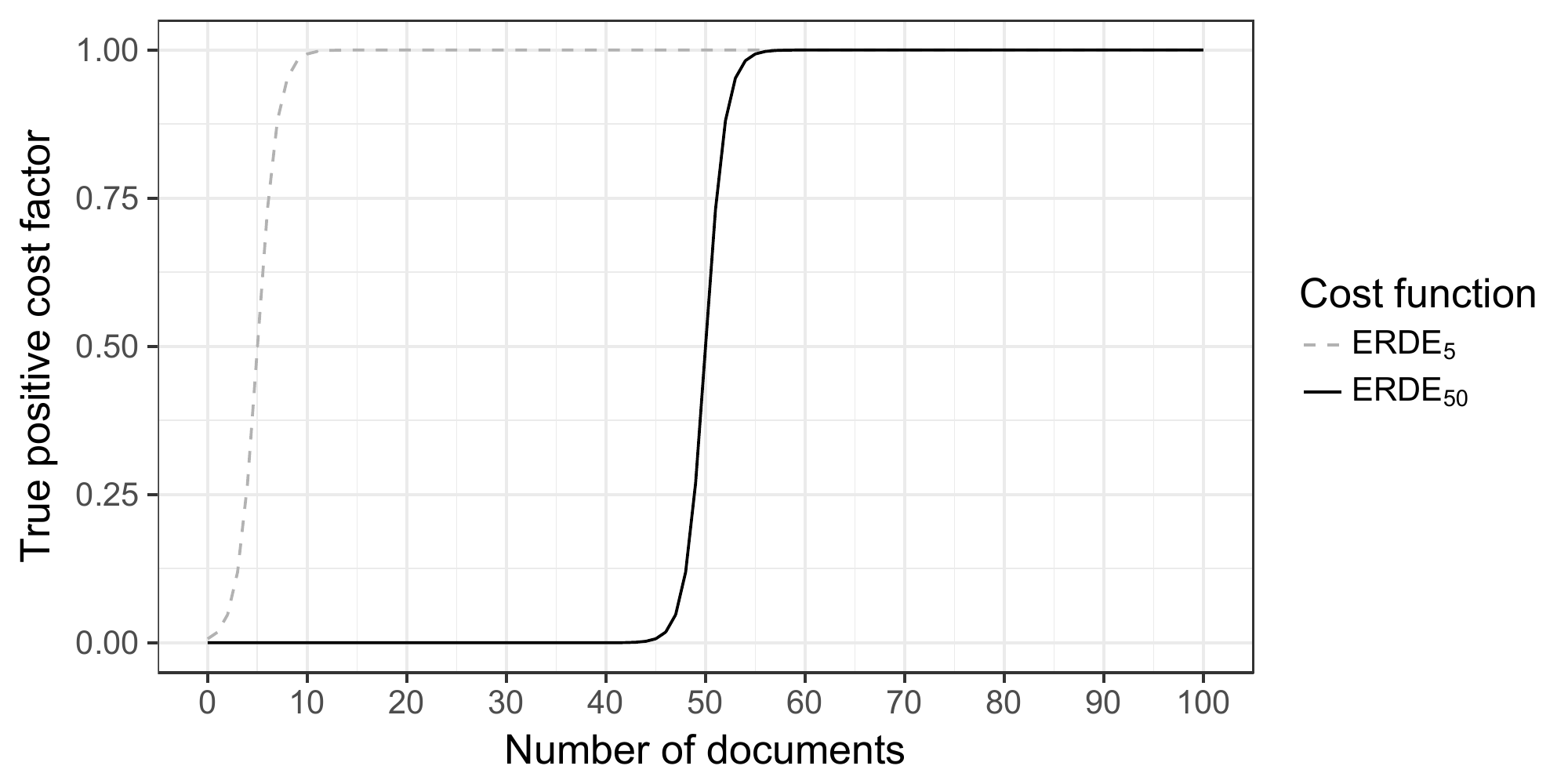}
	\caption{Plot of the true positive cost factor $lc_{o}(k)$ for $ERDE_5$ and $ERDE_{50}$.}
	\label{fig:erde_plot}
\end{figure}

Since the results given for the baseline experiments in the original paper\cite[p. 11]{losada2016test} were obtained by using systems that could submit a prediction after reading each message per user separately, they cannot be compared to results of the actual eRisk task that required to read a whole chunk of between one and 200 messages per user. As Fig. \ref{fig:erde_plot} illustrates, this means that depressed users with about ten and more (for $ERDE_5$) or about 55 and more (for $ERDE_{50}$) messages per chunk basically cannot be predicted correctly because the cost would be very close to $c_{fn}$. Table \ref{tab:erde_perfect} shows the $ERDE_o$ scores of perfect predictions ($F_1 = 1.0$) submitted after reading $n$ chunks with no predictions submitted in the chunks before this one. It also includes the corresponding scores obtained from $ERDE_o^\%$ described at the end of this section. The scores obtained for $n = 1$ are the best possible $ERDE_o$ scores for this task, while $n = 10$ gives the best possible scores for a system that has read all messages. Only predicting the 18 depressed users with less than ten messages per chunk as early as possible and predicting every other user as negative, results in an $F_1$ score of $0.51$ ($1.0$ precision and $0.35$ recall) but still obtains an $ERDE_5$ score of $10.61\%$ and $ERDE_{50}$ of $8.48\%$. The additional $F_1$ score is therefore especially important to evaluate systems in the general task of depression detection. To achieve better $ERDE_o$ scores, systems not only have to be optimized for this task but also need optimized prediction thresholds to make early predictions without too many false positives. This twofold optimization makes this task especially challenging.

\begin{table}[!t]
	\renewcommand{\arraystretch}{1.3}
	\caption{Scores for Perfect Predictions of the eRisk 2017 Test Data After Reading $n$ Chunks According to the Original $ERDE_o$ and the Newly Proposed $ERDE_o^\%$ Score.}
	\label{tab:erde_perfect}
	\centering
	\begin{tabular}{ccc|cc}
		\toprule
		$n$ & $ERDE_5$ & $ERDE_{50}$ & $ERDE_{20}^\%$ & $ERDE_{50}^\%$\\
		\midrule
		1 & 10.60\% & 3.74\% & 0.00\% & 0.00\%\\
		2 & 12.00\% & 5.37\% & 6.48\% & 0.00\%\\
		5 & 12.85\% & 8.47\% & 12.97\% & 6.48\%\\
		10 & 12.97\% & 10.48\% & 12.97\% & 12.97\%\\
		\bottomrule
	\end{tabular}
\end{table}

All experiments described in this paper are based on the exact same training and test data as the eRisk 2017 task and also process it by evaluating the same chunks of test data in chronological order. This ensures that the results are directly comparable to those of the pilot task.

Nevertheless, as the detailed look at the $ERDE_o$ function shows, there is a need to modify this function especially for future work with data in chunks. A first modification of $ERDE$ has already been proposed by Loyola et al.\cite{loyola2017learning} and, in addition to making the score usable for multi-class problems, it mainly consists of an altered cost function defined now as:
\begin{equation}
	lc_{o}(k) = \dfrac{k}{o}\,,
\end{equation}
where $o$ is no longer used to parameterize the cost but equals the number of documents per user and $k$ is the number of documents already read for this user. This ensures that the cost is actually based on the proportion of read documents instead of a fixed number. Still, there is no way to parameterize this function and it immediately grows linearly, without any way to predict a subject correctly with a cost of zero.

We therefore propose a modification of the original sigmoid cost function:
\begin{equation}
	lc_{o}(p) = 1 - \dfrac{1}{1 + e^{p - o}}
\end{equation}
\begin{equation}
	p = \dfrac{100 \cdot k}{n_d}\,,
\end{equation}
where $n_d$ is the total number of documents per user and $k$ still equals to the number of documents already read per user. The cost can still be parameterized by using $o \in [0,100]$ to make the cost grow around the point where $o$ percent of data has been read. This results in a more intuitive cost that grows equally for all users independent of their number of messages. The newly proposed error function based on $lc_{o}(p)$ is denoted as $ERDE_o^\%$ and is evaluated in addition to the original function in Section \ref{chap:experiments}. Table \ref{tab:erde_perfect} shows how this score compares to the original one for perfect predictions of the eRisk 2017 test data. Since at least 10\% of the messages per user have to be read and $o = 18$ is the minimum natural value to achieve an error of 0.00\%, results are shown for $ERDE_{20}^\%$ instead of $ERDE_5^\%$.

Simultaneously to this work, another alternative to $ERDE$ has been proposed by a team that contributed to the eRisk 2017 task as well\cite{sadeque2018measuring}. Their $F_{latency}$ score is based on multiplying the standard $F_1$ score by a factor that is based on the latency of a system, defined as the median number of posts the system has read before predicting the positive cases. In addition, they substitute the sigmoid cost function of $ERDE_o$ by a function that increases more slowly and calculates to a penalty of $0.5$ for the median number of posts in the dataset. Because this score is also tied to the absolute number of read messages, the variance of the available messages per user in the eRisk data would lead to the same problems as described above for $ERDE_o$.

\section{Linguistic Metadata}
\label{chap:metadata}
Augmenting the classification of text sequences with user-level metadata was one of the main ideas in this team's previous work for the eRisk task at CLEF 2017\cite{trotzek2017erisk}. This section builds upon this previously described set of metadata features and is aimed to further describe and extend it. All text-based metadata features are extracted from a concatenation of the \textit{text} and \textit{title} field (see Section \ref{sec:dataset}) of each message, apart from obvious exceptions like the average length of these two fields. All features were calculated separately for each document of a user and then either averaged or summed up as described below.

\subsection{Word and Grammar Usage}
Several features based on counts of specific words or parts of speech (POS) have already been used for this team's work at the eRisk 2017 task and have been examined in the corresponding paper\cite{trotzek2017erisk}. As described in Section \ref{sec:language}, effects of depression on word and grammar usage are well-known and can include, for example, an increased usage of pronouns---especially personal pronouns---, the word ``I'' in particular, and verbs in past tense. Based on these previous findings, occurrences of the word ``I'' were counted separately in the text and title of each message in the dataset. In addition, past tense verbs, personal pronouns, and possessive pronouns were counted in the concatenation of text and title by utilizing the default POS tagger of the Python NLTK framework\footnote{\url{http://www.nltk.org/book/ch05.html}, accessed on 2018-03-13}. As an alternative to this approach, a total of 93 lexicon-based features can be obtained from the LIWC 2015 tool\cite{tausczik2010psychological,pennebaker2003psychological}. Besides features referring to a specific POS as well, this also includes categories like emotions, informal language, or time orientations. LIWC also calculates four summary variables that represent the authenticity, emotional tone, confidence or leadership, and the amount of analytical thinking of a text. Section \ref{sec:meta_features} describes which of these features have been utilized for the experiments of this work. For future work, this approach could likely be enhanced by utilizing more modern POS tagging approaches\cite{huang2015bidirectional, choi2016dynamic} or a POS tagger that was trained specifically on a social media text domain\cite{neunerdt2013part}.
The five final word usage features have also already been used for the participation in eRisk 2017 and aim to count very specific, hand-picked phrases that could be a strong indicator of positive cases. They count the occurrences of the exact phrases ``my depression'', ``my anxiety'', and ``my therapist'' as well as names of some common antidepressants\footnote{\url{https://www.webmd.com/depression/guide/depression-medications-antidepressants}, accessed on 2018-03-14} and variations of the phrase ``I was diagnosed with depression''\footnote{Only including the word ``I'' and an explicit diagnosis as in ``I've been diagnosed with anxiety and depression'' or ``I was diagnosed with major depressive disorder''}. These very explicit features are less aimed at finding early indications of depression but at predicting the obvious cases correctly, which is important for the given task as well. In contrast to the other metadata features, this count is summed up over all documents of a user to make this a strong feature even if only present in few or a single document.

\subsection{Readability}
Measuring the readability or complexity of written text is a well-established idea and various different measures exist, while most of them return a result that corresponds to the school years in the US needed to understand a text. The given dataset cannot really be used as an indicator whether depressed persons write more or less complex texts because of the general difference of text quality between the classes. Since the control subjects were chosen randomly, they often differ drastically from the depressed subjects, who might use reddit to discuss their problems or generally talk to other people. Messages of the control group often simply consist of news headlines, a single short sentence, or even a single word. Readability metrics can therefore help to distinguish messages containing discussions and explanations from such simple content that is unlikely to help with the identification of depression.
Several standard measures for text readability have been calculated for the text content and the four of them with the highest correlation to the class label in the training data have been selected as metadata features, namely \emph{Gunning Fog Index} (FOG)\cite{gunning1952technique}, \emph{Flesch Reading Ease} (FRE)\cite{flesch1948new}, \emph{Linsear Write Formula} (LWF)\footnote{Originally developed by the U.S. Air Force without any publicly available references}\cite{christensen2006readability}, and \emph{New Dale-Chall Readability} (DCR) \cite{dale1948formula,chall1995readability}.

\subsection{Emotions and Sentiment}
As sentiment analysis is focussed on extracting opinions, affects, and emotions from written texts\cite{pang2008opinion}, it seems natural that knowledge from this area can also be very useful to find emotional statements in the field of mental health text classification. Especially the emotions authors express towards their personal situation could be an important indicator. While it would be possible to use the output of any state-of-the-art sentiment classification model\cite{zhang2018deep} as an additional feature, this work has focussed on the use of lexicons to quickly analyze the general helpfulness of sentiment features in this dataset. First of all, the already described LIWC tool includes two features for positive and negative emotions and separate features that indicate anxiety, anger, or sadness. In addition to this, the NRC Emotion Lexicon\cite{mohammad2013emotion} and two general sentiment lexicons, namely the Opinion Lexicon\footnote{\url{https://www.cs.uic.edu/~liub/FBS/sentiment-analysis.html\#lexicon}, accessed on 2018-03-01} and the VADER Sentiment Lexicon\cite{gilbert2014vader}, have been used. There also exist several other lexicons that have not been evaluated, for example from the World Well-Being Project at University of Pennsylvania\footnote{\url{http://www.wwbp.org/lexica.html}, accessed on 2018-03-01}. The NRC Emotion Lexicon contains 14,182 words that can be flagged as positive or negative and as belonging to one or more of the emotions anger, anticipation, disgust, fear, joy, sadness, surprise, and trust. The VADER lexicon includes 7,517 terms (including emoticons) and their mean sentiment value based on the judgement of ten human annotators on a scale between -4 (extremely negative) to 4 (extremely positive). Finally, the Opinion Lexicon consists of two lists with 2,006 positive and 4,783 negative words.
The corresponding counts or scores obtained from these lexicons for the eRisk 2017 dataset were again averaged over all documents of a user. Unfortunately, for this specific dataset no relevant correlation between these features and the class label could be observed. Indeed, the positive (depressed) class contains slightly more emotions and sentiments of all kinds, which might again indicate the general difference of text quality and content between the depressed subjects and the control group. As the emotion and sentiment features were of no use in this specific case, they were not included in the final set of metadata features used in the experiments of this work. Nevertheless, it can be assumed that they would be more meaningful when used with a text corpus that generally included more sentiment and emotion in both classes by using a control group that more closely resembles the target group.

\subsection{Metadata Feature Summary}
\label{sec:meta_features}
Table \ref{tab:metadata} summarizes the 17 metadata features that have been described above excluding the features obtained from LIWC. In addition to these, the ten LIWC features with the highest correlation to the class label in the training data have been selected. Because the LIWC lexicon includes several variations, misspellings, and abbreviations, it is accepted that some of these features already occur in the previously described feature set and therefore introduce a slight redundancy. The selected LIWC features are the number of function words, variations of the word ``I'' (e.g. including abbreviations containing ``I'' as well as ``me'' and ``myself''), all pronouns, personal pronouns, verbs, words indicating a cognitive process, words with a focus on the present, the total number of lexicon words found, and the two calculated summary variables indicating analytical thinking and authenticity. To build the user-level metadata vector for the experiments described in Section \ref{chap:experiments}, these features were again averaged over all documents of the same user.
The described metadata features result in a 27-dimensional vector per user. The concatenated feature vectors of all users are standardized as described in Section \ref{sec:experiment_setup} before being used as input to a classifier. The five counts of specific terms are transformed into boolean flags by representing a value above 0 as 1 and a value of 0 as -1, similar to the previous work using LSTM networks\cite{trotzek2017erisk}. As the experiments will show, this set of metadata features alone can lead to very good results on the eRisk 2017 dataset.

\begin{table}[!t]
	\renewcommand{\arraystretch}{1.3}
	\setlength{\tabcolsep}{1mm}
	\caption{User Based Metadata Features Used in Combination with the Selected LIWC Features.}
	\label{tab:metadata}
	\centering
	\begin{tabular}{lll}
		\toprule
		Feature & Type & Description\\
		\midrule
		``I'' in the text & avg. & only the word ``I''\\
		``I'' in the title & avg. & \ditto{3.6} \\
		Possessive pronouns & avg. & based on POS tagging\\
		Personal pronouns & avg. & \ditto{4.6} \\
		Past tense verbs & avg. & \ditto{4.6} \\
		4 readability scores & avg. & FOG, FRE, LWF, and DCR\\
		Month of the writings & avg. & \\
		Text length & avg. & \\
		Title length & avg. & \\
		Depression & sum & ``my depression''\\
		Anxiety & sum & ``my anxiety''\\
		Therapist & sum & ``my therapist''\\
		Diagnosis & sum & e.g. ``I was diagnosed with depression''\\
		Antidepressants & sum & e.g. ``zoloft'' and ``paxil''\\
		\bottomrule
	\end{tabular}
\end{table}

\section{Neural Network Models}
\label{chap:network}
The following sections are used to describe the neural network models that were used in the experiments of this work. All of these models are based on a document vectorization using neural word embeddings. The general concept of word embeddings and the specific models utilized in this case is described in Section \ref{sec:embeddings}. Afterwards, the following section is used to explain the type of network and the model architecture that was implemented for the experiments.

\subsection{Word Embeddings}
\label{sec:embeddings}
Neural word embeddings have become a popular and efficient way to model words and interactions between them for purposes like text classification tasks. They date back to the concept of distributed word representations\cite{hinton1986distributed} that, in contrast to local representations, do not handle each word separately with a single neuron, but use several neurons to represent a word and let each neuron be part of the description of several words. This enables distributed representations to learn general concepts of language instead of just independent word representations. One of the most important and still popular methods to train word embeddings was published by Google as word2vec\cite{mikolov2013efficient,mikolov2013distributed}, which consists of two neural network architectures---namely the Continuous Bag of Words (CBoW) and the (Continuous) Skip-gram (SG) architecture. The concept of word2vec was developed further by Facebook and published as fastText in 2017\cite{joulin2017bag,bojanowski2017enriching,mikolov2017advances}, which also directly offers text classification. While being based on the same two model architectures as word2vec, fastText represents words as bags of character $n$-grams and thus allows to obtain vectorizations even for unknown words. A different approach to learn word embeddings, GloVe, has been published by researchers of the Stanford NLP group\cite{pennington2014glove}. GloVe aims to combine the advantages of local context window approaches like Skip-gram with those of global matrix factorization models like Latent Semantic Analysis (LSA).

Pre-trained word vectors obtained from large corpora are available for both fastText\footnote{\url{https://fasttext.cc/docs/en/pretrained-vectors.html} and \url{https://fasttext.cc/docs/en/english-vectors.html}, accessed on 2018-03-07} and GloVe\footnote{\url{http://nlp.stanford.edu/projects/glove}, accessed on 2018-03-07}. For this work, a 50-dimensional GloVe model trained on Wikipedia and the Gigaword 5 news corpus as well as a 300-dimensional GloVe model trained on Common Crawl were chosen. In addition, three 300-dimensional pre-trained fastText models based on similar corpora were used.

\begin{table*}[!t]
	\renewcommand{\arraystretch}{1.3}
	\caption{Characteristics of the GloVe and fastText Word Embeddings Used in this Paper.}
	\label{tab:embeddings}
	\centering
	\begin{tabular}{lc>{\centering}p{1.9cm}>{\centering}p{1.7cm}c}
		\toprule
		Model & Dimension & Corpus tokens (in billion) & Word vectors (in milllion) & Words of eRisk 2017\\
		\midrule
		GloVe Wiki + News & 50 & 6 & 0.4 & 81.9\%\\ 
		GloVe Crawl & 300 & 42 & 2 & 93.2\%\\ 
		fastText Wiki & 300 & ? & 2.5 & 81.7\%\\ 
		fastText Wiki + News & 300 & 16 & 1 & 79.5\%\\ 
		fastText Crawl & 300 & 600 & 2 & 88.5\%\\ 
		\midrule
		fastText reddit & 300 & 49.9 & 6 & 99.7\%\\ 
		\bottomrule
	\end{tabular}
\end{table*}

Finally, to also examine word embeddings that better fit the domain of reddit messages (or social media platforms in general), an own fastText model was trained on a dataset containing all reddit comments between October 2007 and May 2015\footnote{\url{https://www.reddit.com/r/datasets/comments/3bxlg7/i_have_every_publicly_available_reddit_comment/}, accessed on\\2018-03-07}. The total dataset consists of about 1.7 billion messages that we preprocessed and tokenized in a way that preserves emoticons, punctuation, and words that include special characters (e.g. censored words). The preprocessing step also included to replace any references to reddit users (in the form of /u/$<$username$>$) by a generic phrase ``ref\_user'' to prevent any connections to actual users in the resulting word embeddings. Similarly, any reference to a subreddit (in the form of /r/$<$subreddit$>$) was replaced by the phrase ``ref\_subreddit\_$<$subreddit$>$'' to be able to learn a vector representation of them as well that can be regarded as their topic. No stemming or stopword removal of any kind was done. The resulting tokens of each message were lowercased (with the exception of emoticons) and separated with a space to enable fastText to properly treat punctuation\footnote{\url{https://github.com/facebookresearch/fastText/issues/333}, accessed on 2018-03-07}. Since the dataset also contains messages written in different languages than English and a sophisticated language detection classifier would have required too much time for so many documents, a simple language detection based on stopword counts was utilized: Only messages with more English stopwords than ones from other languages\footnote{based on the 2,400 stopwords for 11 languages of the NLTK Python package, \url{http://www.nltk.org/book/ch02.html#annotated-text-corpora}, accessed on 2018-03-07} were retained (thus also discarding messages without any stopwords). This resulted in a final training corpus of 1.37 billion messages and a total of 49.9 billion tokens. The C++ implementation of fastText was used to train a 300-dimensional CBoW model without subword information and default values for all other parameters, which contains the 6 million unique tokens that occur at least five times in this corpus. Training took about 26 hours using 12 threads on an Intel Xeon E5-2687W 3.10GHz and needed about 17GB of RAM.

The characteristics of all utilized word embeddings are shown in Table \ref{tab:embeddings}. It also contains the amount of words each model includes of the 85,558 words that occur in writings of at least two users of the eRisk 2017 dataset after the same preprocessing and tokenization steps as described above. This set of words is used in the experiments described in Section \ref{chap:experiments}.

\begin{figure*}[!t]
	\centering
	\includegraphics[width=7.16in]{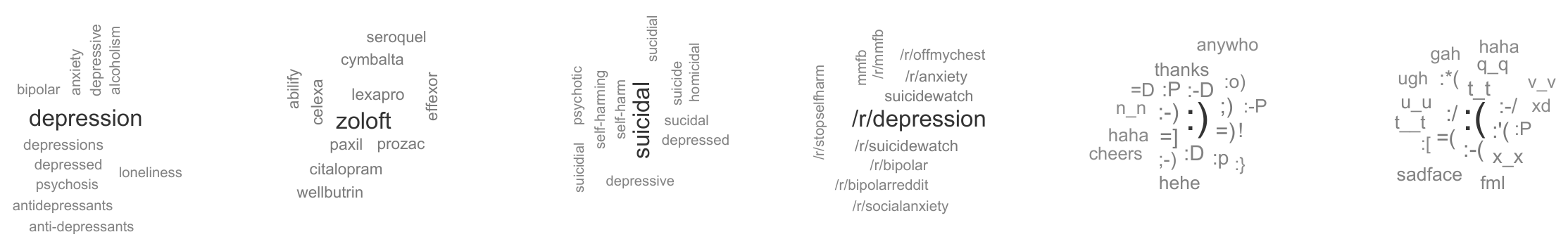}
	\caption{Nearest neighbors for six hand-picked tokens in the fastText model trained on reddit comments according to cosine similarity.}
	\label{fig:token_nn}
\end{figure*}

As a qualitative analysis of the self-trained fastText model, it is possible to examine the nearest neighbors of some hand-picked exemplary tokens according to cosine similarity. Fig. \ref{fig:token_nn} displays six word clouds with the corresponding example token in the middle and its ten (20 in the case of emoticons) nearest neighbors around it. These examples especially illustrate that the model trained on reddit messages is indeed able to identify similar emoticons and subreddits, which is both not possible using the pre-trained fastText or GloVe models. The closest neighbors of the depression subreddit also include terms like ``suicidewatch'' and ``mmfb'' (make me feel better), illustrating relations that were learnt to terms besides other subreddits. It has also learnt similar embeddings for antidepressants like Zoloft. While this as well as similar words to the terms ``depression'' and ``suicidal'' can also be observed using the pre-trained models, their nearest neighbors seem slightly more medical and from a more neutral perspective (e.g. ``ssri'' or ``sertraline'' close to ``zoloft'', ``melancholia'' or ``insomnia'' for ``depression'', and ``deranged'' or ``delusional'' for ``suicidal'') than those of the model trained on reddit. Also, especially the fastText Crawl model returns neighboring terms like ``depression.This'' or ``depression.What'', which might indicate a preprocessing problem concerning punctuation.

To further investigate the extent to which the self-trained fastText reddit model has learnt a general model of the English language, the standard word analogy dataset\cite{mikolov2013efficient} can be used as one indicator. Table \ref{tab:analogy} compares results on the word analogy dataset published for state-of-the-art models to the results of the new model trained on reddit messages. The dataset was originally created to evaluate the word embeddings of word2vec and consists of 8.869 semantic as well as 10.675 syntactic analogy questions: Given an analogy (e.g. Athens and Greece) and a third word (e.g. Oslo), word embeddings have to return the fourth word (Norway in this case) as closest vector to the result of $vec(``Greece'') - vec(``Athens'') + vec(``Oslo'')$ according to cosine similarity. While the results are far from the ones obtained by the state-of-the-art fastText models trained on Wikipedia and news articles, especially the result in the syntactic category illustrates that even the training on these much less formal documents has lead to a decent model of the English language.
\begin{table}[!t]
	\renewcommand{\arraystretch}{1.3}
	\caption{Semantic, Syntactic, and Total Accuracies on the Word Analogy Dataset for 300-Dimensional Word Embeddings}
	\label{tab:analogy}
	\centering
	\begin{tabular}{lccc}
		\toprule
		Model & Sem. & Syn. & Total\\
		\midrule
		word2vec\cite{mikolov2013distributed} & 63 & 58 & 61\\
		GloVe Crawl\cite{pennington2014glove} & 82 & 69 & 75\\
		fastText Wiki\cite{bojanowski2017enriching} & 78 & 75 & 76\\
		fastText Wiki + News\cite{mikolov2017advances} & \textbf{90} & \textbf{84} & \textbf{87}\\
		fastText Crawl\cite{mikolov2017advances} & 87 & 82 & 85\\
		\midrule
		fastText reddit & 56 & 74 & 66\\
		\bottomrule
	\end{tabular}
\end{table}

\subsection{Neural Network Architecture}
\begin{figure*}[!t]
	\centering
	\includegraphics[width=7.16in]{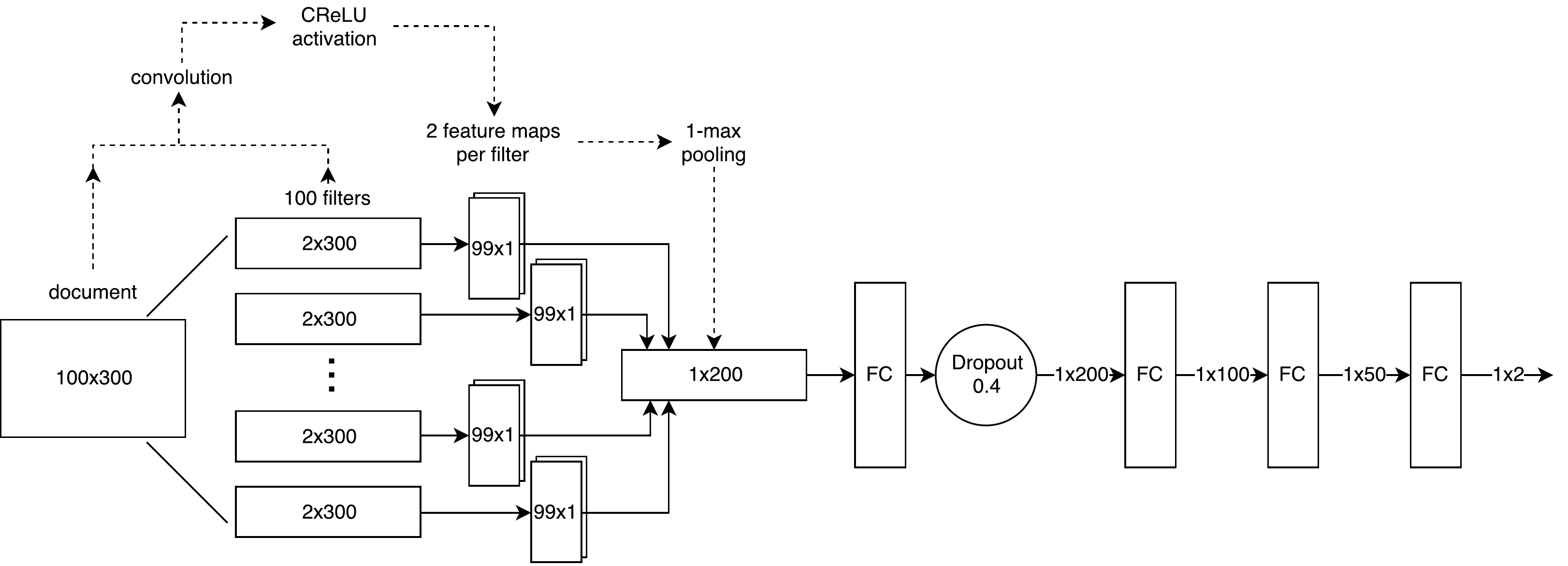}
	\caption{Network architecture used for the convolutional neural network models described in this paper. Adapted from \cite{zhang2017sensitivity}.}
	\label{fig:cnn_architecture}
\end{figure*}

Convolutional Neural Networks (CNN)\cite{lecun1989generalization} have been utilized to achieve outstanding results especially in the area of image classification and are generally viable for data with a grid-like structure\cite{goodfellow2016deep}. Recently, studies have shown that they can also be used effectively for text classification tasks\cite{zhang2017sensitivity}. Fig. \ref{fig:cnn_architecture} displays the architecture of the simple CNN used for the experiments described in this paper, which is based on the one-layer CNN for sentence classification described by Zhang and Wallace\cite{zhang2017sensitivity}. Similar to this sentence classification network, it consists of only a single CNN layer but uses a total of 100 filters with a height of 2 and a width corresponding to the word vector dimensions. Concatenated Rectified Linear Units (CReLU)\cite{shang2016understanding} are used as activation function for the convolutional layer as well as for the fully-connected layers, resulting in twice as many outputs due to the concatenation with the negated activation. 1-max pooling is used to obtain a single scalar from each filter, which results in a 200-dimensional vector due to the CReLU activation. This output is then propagated through three fully-connected layers with, again, CReLU activation, of which the first one applies dropout to its output. The fourth and final layer applies softmax to the output.
As input, the network receives each document of a user individually in the form of the 100 first word vectors per document, while using zero-padding for documents with less than 100 words. This results in a $100\times 50$ or $100\times 300$ dimensional input matrix depending on the used word embedding as described in the previous section. The limitation to 100 words (or even less) is possible as the number of words per document ranges between 1 (when ignoring the empty documents) and 6,487 but has a mean of 34.58 according to the tokenization done for this work. As this results in a separate output for each document per user, the 98th percentile of these outputs is used as final prediction for the user. This value is chosen instead of the mean to give more weight to documents with a higher probability.

In addition to the described model architecture, it has been tried to directly use the metadata features as a second input to the network. An approach similar the the previous one at eRisk 2017\cite{trotzek2017erisk}, where the metadata features were fed through a fully-connected layer and then concatenated with the output of an LSTM layer, did not lead to better results than just using the text input. The same applies to the idea to use the final $1\times 50$ dimensional vector before the softmax layer as additional input for a metadata classifier like a logistic regression. Since the results of networks using the metadata differed only marginally from the text-only network described above, only results of the latter will be reported in the following to reduce the complexity. Future work will be needed to explore better ways to directly merge the CNN output with the metadata. This could, for example, be done by implementing a dedicated fusion component into the neural network, similar to work done for gender identification based on texts and images at the CLEF 2018 PAN workshop\cite{takahashi2018pan}. In this work, a simple late fusion ensemble will show that the best results so far can indeed be achieved by combining these features.

\section{Experiments}
\label{chap:experiments}
This section is used to describe the experiments done based on the convolutional neural network and the metadata features as well as their results. Results are compared to the best published results during the eRisk 2017 task as well as other results obtained after the ground truth was released. The scores of each model are reported according to $ERDE_5$, $ERDE_{50}$, and $F_1$, which are the official scores of this task, and also based on the newly proposed $ERDE_{20}^\%$, $ERDE_{50}^\%$, and $F_{latency}$.

\subsection{Experiment Setup}
\label{sec:experiment_setup}
For the experiments conducted during this work, the same process used during the eRisk 2017 task was reproduced: The available test documents were processed in the same ten chunks that contain 10\% of the writings obtained from each user. Training is done once on the full training dataset. Afterwards, test chunks are processed in sequential order, while the documents of the previous chunks are always used again. The only exception to this process is the model called ``Meta LR Wait'' in the following evaluation section, which is a logistic regression based on metadata features that was configured to only submit a prediction after the final chunk. Similar ``waiting'' models were also utilized by some teams during eRisk 2017 and can be interesting to evaluate the possible $F_1$ score, while neglecting the early detection aspect and therefore the $ERDE_o$ scores. Since the models based on metadata use features averaged over all documents of the same user, they were also calculated for each chunk separately, again using documents from earlier chunks as well.
An additional parameter for the early detection models is the prediction threshold that determines whether a model is confident enough to predict a subject as positive (depressed) or whether it waits for more data. While these thresholds were based on cross-validation using the training data for this team's participation in the eRisk 2017 task and included the number of documents already read in multiple threshold levels\cite{trotzek2017erisk}, the experiments in this work are based on a single threshold value that achieved the best test result for the specific model. This is likely to lead to an overfitting on the specific test data but also allows to compare the best possible results of the utilized models. Generally, prediction thresholds between 0.5 and 0.7 lead to a balanced result in all scores, while higher thresholds can often maximize $ERDE_5$ but severely decrease $F_1$. This fits the observations described in Section \ref{sec:task}: Since the correct prediction of so few depressed test subjects actually has an effect on $ERDE_5$, it is often best to submit fewer predictions overall and therefore simply minimize false positives. Negative (non-depressed) predictions were only submitted after seeing the final chunk.
The models based on user metadata features all utilized the same logistic regression classifier. The 27 features described in Section \ref{sec:meta_features} were first standardized to have a mean of 0 and unit variance, with exception of the boolean flag features that already have a value of either -1 or 1. The resulting scaled feature vector was then used to train a logistic regression classifier and later predict probabilities for the test subjects of each chunk.

\subsection{Evaluation}
Table \ref{tab:result_comparison} displays the results achieved in this work in comparison to previously published results for the same dataset and task. The first three rows in this table represent the best results during the eRisk 2017 task and are therefore solely optimized based on cross-validation over the training data, while the next two results have been achieved after the ground truth was published. All results after these have been achieved as part of this work. The models corresponding to the name of a word embedding refer to a CNN using this embedding as input vectorization, the models named ``Meta LR'' refer to the logistic regression based on metadata, and the final four results were obtained by calculating the mean of the metadata probabilities and the neural network output. Although these outputs have not been calibrated (e.g. by using Platt scaling\cite{platt1999probabilistic}), this simple late fusion ensemble lead to the best achieved $ERDE_o$ scores and recall. As expected, the best overall $F_1$ score could be obtained by waiting for the last chunk and only then submitting predictions based on the metadata LR. Interestingly, this model would still have achieved the seventh best $ERDE_{50}$ score in the eRisk 2017 task out of 30 submissions, which again illustrates how difficult $ERDE_o$ is to interpret because it is based on the absolute number of documents.
The prediction thresholds have been chosen to represent the best possible $ERDE_o$ scores that still include a viable $F_1$ score. A second threshold has been reported for the self-trained fastText reddit model and the metadata LR to illustrate to which extent slightly different thresholds can have an effect on $ERDE_o$ scores. As already described, especially optimizing $ERDE_5$ often includes impairing $F_1$ score. The reported results contain the best scores published for this task so far and, importantly, achieve a balanced result among all scores. Although $ERDE_o$ and $F_1$ are difficult to maximize simultaneously, the results show that the described models are able to do so.

\begin{table}[!t]
	\renewcommand{\arraystretch}{1.3}
	\setlength{\tabcolsep}{1.5mm}
	\caption{$ERDE_o$ Scores, $F_1$ Score, Precision, and Recall of the Utilized Models and Previous Publications. The Second Column Displays the Prediction Threshold.}
	\label{tab:result_comparison}
	\centering
	\begin{tabular}{lc|ccccc}
		\toprule
		Model & $p >$ & $ERDE_5$ & $ERDE_{50}$ & $F_1$ & P & R\\
		\midrule
		UNSLA\cite{losada2017erisk} & & 13.66 & 9.68 & 0.59 & 0.48 & 0.79\\
		FHDO-BCSGA\cite{losada2017erisk} & & 12.82 & 9.69 & 0.64 & 0.61 & 0.64\\
		FHDO-BCSGB\cite{losada2017erisk} & & 12.70 & 10.39 & 0.55 & 0.69 & 0.46\\
		\midrule
		TVT-NB\cite{villegas2017erisk} & & 13.13 & 8.17 & 0.54 & 0.42 & 0.73\\
		TVT-RF\cite{villegas2017erisk} & &12.30 & 8.95 & 0.56 & 0.54 & 0.58\\
		\midrule
		GloVe W+N & 0.5 & 12.95 & 7.57 & 0.63 & 0.56 & 0.73\\
		GloVe Crawl & 0.7 & 12.98 & 8.59 & 0.63 & 0.58 & 0.69\\
		fastText Wiki & 0.6 & 13.06 & 8.17 & 0.57 & 0.47 & 0.71\\
		fastText W+N & 0.55 & 13.11 & 7.95 & 0.60 & 0.49 & 0.77\\
		fastText Crawl & 0.6 & 13.01 & 8.60 & 0.64 & 0.60 & 0.67\\
		fastText reddit & 0.7 & 13.52 & 8.04 & 0.62 & 0.51 & 0.79\\
		fastText reddit & 0.8 & 12.71 & 9.23 & 0.56 & 0.63 & 0.50\\
		Meta LR & 0.35 & 12.65 & 8.57 & 0.66 & 0.59 & 0.73\\
		Meta LR & 0.55 & 12.35 & 9.86 & 0.65 & 0.72 & 0.60\\
		Meta LR Wait & 0.35 & 13.32 & 11.33 & \textbf{0.73} & \textbf{0.77} & 0.69\\
		\midrule
		G W+N + Meta LR & 0.45 & 12.34 & 8.93 & 0.71 & 0.72 & 0.69\\
		fT Wiki + Meta LR & 0.35 & 13.52 & \textbf{7.29} & 0.55 & 0.41 & \textbf{0.85}\\
		fT Wiki + Meta LR & 0.5 & \textbf{12.13} & 8.77 & 0.71 & 0.71 & 0.71\\
		fT reddit + Meta LR & 0.55 & 12.46 & 8.77 & 0.67 & 0.69 & 0.65\\
		\bottomrule
	\end{tabular}
\end{table}

In addition to comparing the achieved results to previously published results for this task, Table \ref{tab:result_erdep} shows how the same models with the same prediction thresholds would have scored according to the newly proposed $ERDE_o^\%$ score as well as the $F_{latency}$ score from \cite{sadeque2018measuring}. While the ``Meta LR Wait'' model is now scored equally bad in both criteria because it had to read all documents, the CNN scores now tend to be better than the ones obtained for the metadata models alone. Still, the best overall $ERDE_{o}^\%$ scores could be achieved by the same ensemble. The additional models with higher thresholds that were previously included to obtain a better $ERDE_5$ score (namely fastText reddit with $p > 0.8$ and Meta LR with $p > 0.55$) now result in the worst overall $ERDE_o^\%$ scores next to the waiting model. This again indicates that especially optimizations of $ERDE_5$ do not necessarily mean a better classification result. The 50-dimensional GloVe model achieves the best $F_{latency}$ score, which is also better than the best score reported in the original paper (0.389) for the same dataset\cite{sadeque2018measuring}.

\begin{table}[!t]
	\renewcommand{\arraystretch}{1.3}
	\caption{Scores of the Above Models According to the Proposed $ERDE_o^\%$ Criterium and the $F_{latency}$ Measure from \cite{sadeque2018measuring}.}
	\label{tab:result_erdep}
	\centering
	\begin{tabular}{lc|cccc}
		\toprule
		Model & $p >$ & $ERDE_{20}^\%$ & $ERDE_{50}^\%$ & $F_{latency}$\\
		\midrule
		GloVe W+N & 0.5 & 8.70 & 7.08 & \textbf{0.52}\\ 
		GloVe Crawl & 0.7 & 9.44 & 6.58 & 0.44\\ 
		fastText Wiki & 0.6 & 8.18 & 6.69 & 0.46\\ 
		fastText W+N & 0.55 & 8.34 & 6.10 & 0.49\\ 
		fastText Crawl & 0.6 & 9.60 & 7.23 & 0.48\\ 
		fastText reddit & 0.7 & 8.53 & 5.53 & 0.51\\ 
		fastText reddit & 0.8 & 10.21 & 9.21 & 0.35\\ 
		Meta LR & 0.35 & 9.32 & 7.32 & 0.43\\
		Meta LR & 0.55 & 10.74 & 8.74 & 0.40\\
		Meta LR Wait & 0.35 & 13.32 & 13.32 & 0.26\\
		\midrule
		G W+N + Meta LR & 0.45 & 9.68 & 7.56 & 0.45\\
		fT Wiki + Meta LR & 0.35 & \textbf{6.40} & \textbf{4.78} & 0.49\\
		fT Wiki + Meta LR & 0.5 & 9.21 & 7.47 & 0.48\\
		fT reddit + Meta LR & 0.55 & 9.46 & 7.47 & 0.45\\
		\bottomrule
	\end{tabular}
\end{table}

\section{Conclusion}
\label{chap:conclusion}
This work has been used to examine the currently popular $ERDE_o$ metric for early detection tasks in detail and has shown that especially $ERDE_5$ is not a meaningful metric for the described shared task. Only the correct prediction of few positive samples has an effect on this score and the best results can therefore often be obtained by only minimizing false positives. A modification of this metric, namely $ERDE_o^\%$, has been proposed that is better interpretable in the case of shared tasks that require information to be read in chunks. Exemplary scores using this score have been shown in comparison to $ERDE_o$ scores for the experiments in this work.

Previous experiments for the eRisk 2017 task for early detection of depression have been continued by examining additional user-level metadata features and evaluating a convolutional neural network as text-based depression classifier. State-of-the-art results have been reported for the eRisk 2017 dataset using these two approaches. A new fastText word embedding has been trained on a large corpus of reddit comments. The analysis of the resulting word vectors has shown that the model has learnt some features specific to this domain and is viable for general syntactic questions in the English language as shown based on the standard word analogy task.

As the results presented in this paper are optimized to obtain the best performance on the eRisk 2017 task for comparison to previously published results and among these models, future work will have to show how these models perform on yet unseen data. This has first been done during the eRisk 2018 task\cite{10.1007/978-3-319-98932-7_30}, which used the old dataset as training data and contained 820 new test subjects. In addition, eRisk 2018 contained an additional task aimed at the early detection of anorexia that this team has also participated in. The five submitted predictions achieved the best $F_1$ and $ERDE_{50}$ scores in both tasks and the CNN without metadata in particular achieved the best results in the new anorexia task\cite{trotzek2018erisk}. The same working notes paper for this second participation has also been used to evaluate the modified $ERDE_{o}^\%$ metric for all participants and again shows how especially the original $ERDE_5$ metric favors systems that correctly predict test users with only few documents in total regardless of their overall performance.

As the detailed look at the current $ERDE_o$ metric has shown, one priority of future work in this area should be to agree on a new metric for early detection tasks like eRisk. Ethical issues in this area of research have been reviewed and should find more attention as well. Possibilities to publish the fastText model trained on reddit comments still have to be examined. Concerning the models presented in this work, additional experiments will be necessary to find better ways to integrate the metadata features directly into the neural network. On the other hand, utilizing ensembles of more than just two models and calibrating the resulting probabilities seems promising. Combining word embeddings of two models in a single neural network has also not been evaluated yet. Another possible improvement would be to use recently published language modeling methods like BERT as input for the network and to compare a self-trained model using this approach to the fastText word embeddings of this work.

\ifCLASSOPTIONcompsoc
  \section*{Acknowledgments}
\else
  \section*{Acknowledgment}
\fi

The work of Sven Koitka was partially funded by a PhD grant from University of Applied Sciences and Arts Dortmund, Germany.

\ifCLASSOPTIONcaptionsoff
  \newpage
\fi



\bibliographystyle{IEEEtran}
\bibliography{journal_paper}
%

%
%

%

\begin{IEEEbiography}[{\includegraphics[width=1in,height=1.25in,clip,keepaspectratio]{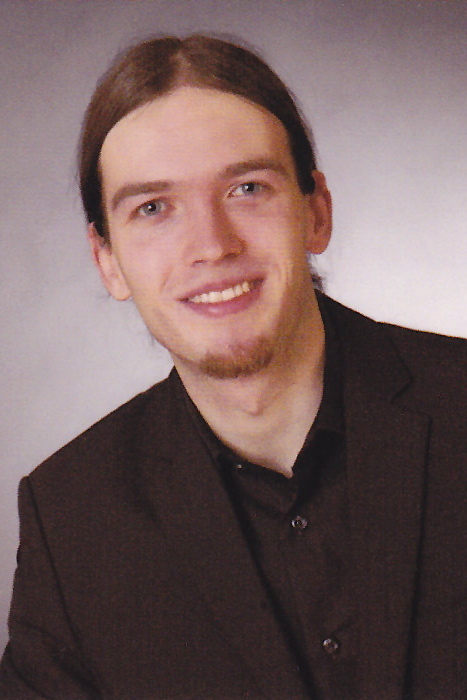}}]{Marcel Trotzek}
received the M.Sc. degree in Computer Science from University of Applied Sciences and Arts Dortmund, Germany, in 2018. His research interests are in the area of machine learning with a focus on text classification.
\end{IEEEbiography}

\begin{IEEEbiography}[{\includegraphics[width=1in,height=1.25in,clip,keepaspectratio]{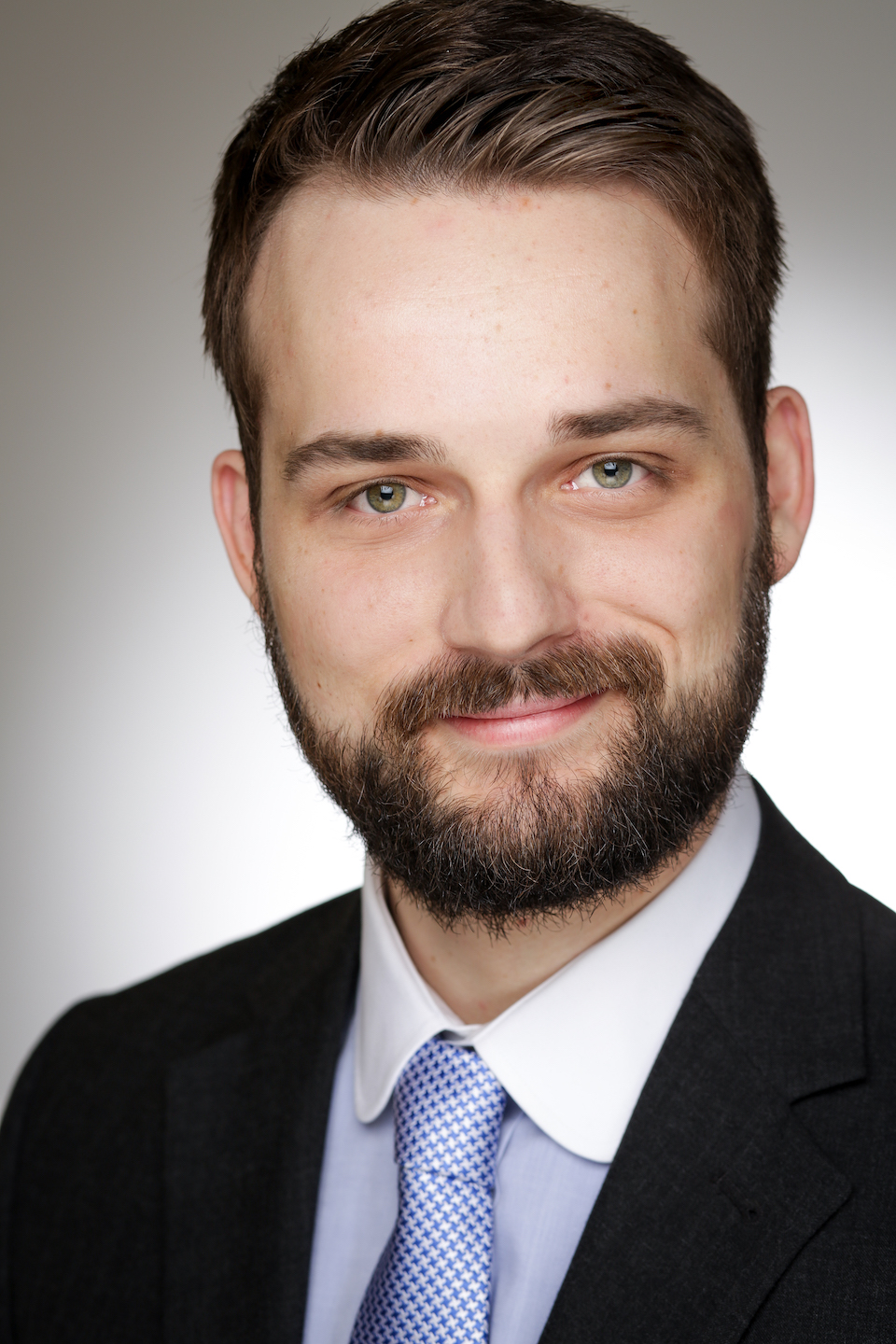}}]{Sven Koitka}
received the M.Sc. degree in Computer Science from the University of Applied Sciences and Arts Dortmund, Germany, in 2016 and is currently working on his Ph.D. degree. His research interests are in the area of machine learning, in particular neural networks with a focus on computer vision.
\end{IEEEbiography}


\begin{IEEEbiography}[{\includegraphics[width=1in,height=1.25in,clip,keepaspectratio]{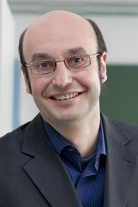}}]{Christoph M. Friedrich}
received a diploma in computer science from the University of Dortmund, Germany (1996) and a Ph.D. in Life Science Engineering from University of Witten/Herdecke, Germany (2006). He joined the computer science department of University of Applied Sciences and Arts Dortmund, Germany, as a Professor of biomedical computer science in 2010. From 2005 to 2010 he led the data mining group at the bioinformatics department of Fraunhofer Institute for algorithms and scientific computing (SCAI). Besides his academic career, he has several years of industrial experience. He is author or co-author of over 80 publications in international journals or conference proceedings. His main field of interest is in machine learning, text mining, biomedical applications of computer vision and bioinformatics.
\end{IEEEbiography}




\end{document}